\documentclass[conference]{IEEEtran}
\usepackage{cite}
\usepackage{amsmath,amssymb,amsfonts}
\usepackage{algorithmic}
\usepackage{graphicx}
\usepackage{textcomp}
\usepackage{xcolor}
\usepackage[export]{adjustbox}
\usepackage{caption}
\usepackage{subcaption}
\usepackage{hyperref}
\usepackage[mathlines]{lineno}
\def\BibTeX{{\rm B\kern-.05em{\sc i\kern-.025em b}\kern-.08em
    T\kern-.1667em\lower.7ex\hbox{E}\kern-.125emX}}
\begin{document}
\title{Mask Guided Gated Convolution for Amodal Content Completion}

\makeatletter 
\newcommand{\linebreakand}{%
  \end{@IEEEauthorhalign}
  \hfill\mbox{}\par
  \mbox{}\hfill\begin{@IEEEauthorhalign}
}
\makeatother 

\author{
\IEEEauthorblockN{Kaziwa Saleh}
\IEEEauthorblockA{\textit{Doctoral School of Applied Informatics}\\\textit{and Applied Mathematics, Óbuda University} \\
Budapest, Hungary \\
kaziwa.saleh@nik.uni-obuda.hu}

\and

\IEEEauthorblockN{S\'{a}ndor Sz\'{e}n\'{a}si}
\IEEEauthorblockA{\textit{John von Neumann Faculty of}\\
\textit{Informatics, \'{O}buda University}\\
Budapest, Hungary \\
szenasi.sandor@nik.uni-obuda.hu}

\and

\IEEEauthorblockN{Zolt\'{a}n V\'{a}mossy}
\IEEEauthorblockA{\textit{John von Neumann Faculty of}\\
\textit{Informatics, \'{O}buda University}\\
Budapest, Hungary \\
vamossy.zoltan@nik.uni-obuda.hu}
}

\maketitle
\begin{abstract}
We present a model to reconstruct partially visible objects. The model takes a mask as an input, which we call weighted mask. The mask is utilized by gated convolutions to assign more weight to the visible pixels of the occluded instance compared to the background, while ignoring the features of the invisible pixels. By drawing more attention from the visible region, our model can predict the invisible patch more effectively than the baseline models, especially in instances with uniform texture. The model is trained on COCOA dataset and two subsets of it in a self-supervised manner. The results demonstrate that our model generates higher quality and more texture-rich outputs compared to baseline models. Code is available at: \url{https://github.com/KaziwaSaleh/mask-guided}
\end{abstract}


\section{Introduction}
Our environment is cluttered and crowded. Objects are often occluded by other object(s), and occlusion can happen in various rates, and positions. Regardless of the fact that we encounter more occluded objects than fully visible ones, we seamlessly and effortlessly can imagine the full shape and gestalt of partially visible objects. This process is called amodal completion. While this task is natural and innate for humans, it is challenging for machines. However, it is essential for many real-world computer vision based applications, such as semantic scene understanding, robotics, autonomous vehicles, and security and surveillance systems. \cite{saleh2021occlusion}.

In order to enable amodal completion in computers, we need three sub-tasks: amodal segmentation, order recovery, and amodal content completion \cite{ao2023image}. Amodal segmentation predicts the whole shape of the object including the missing region, also called amodal mask. The depth order and the occluder-occludee relationship between objects is determined through order recovery. The appearance and RGB content of the occluded area is synthesized via amodal content completion.

Several works are focused on the problem of amodal segmentation \cite{zhu2017semantic,bs3,bs7,bs13,bs16,bs21,bs22,ozguroglu2024pix2gestalt}. However, the problem of amodal content completion remains comparatively less discovered. One of the challenges of amodal content completion is the lack of a training dataset. The existing datasets do not contain ground-truth RGB content of the hidden area of partially visible objects.  

Although amodal content completion may seem conceptually similar to inpainting, they are fundamentally different. While the aim of inpainting is to complete the missing region(s) of an image with no knowledge about the boundary and the structure of the individual objects included in the completion process, amodal completion uses the amodal mask of the objects to guide the reconstruction process. 

The existing models cannot fill the content of the missing part without any traces from the occluder(s). Following the intuition that regions in an object are semantically similar, we attempt to improve the quality of the generated patch through utilizing the visible region of the object to guide the completion process. 

In this paper, we propose a model for amodal content completion that applies a network of gated convolution and contextual attention layers to the input image. We harness the attention from the visible pixels of the occluded object through a mask, which we call weighted mask, that encourages the model to concentrate on the global features of the image and the visible region of the object concurrently, but with more weight given to the latter. The model is optimized through the combination of hinge loss, perceptual loss, pixel-wise $\mathcal{L}$1 loss, style loss, and patch loss.

In order to overcome the lack of training data, we follow the self-supervised approach proposed by Zhan et al. \cite{zhan2020self}. We randomly select an object from the dataset and position it on another object to create occlusion. The amodal segmentation mask for training can then be obtained by merging the modal mask (visible segmentation mask) of the occluded object and the mask for the occluded region. The model is then used to complete the appearance of the occluded area indicated by the mask.

We report the results of our model on COCOA \cite{zhu2017semantic} dataset and two subsets of it. The results show that our model can generate richer semantic and texture details when completing the partially visible objects compared to the baseline models. This illustrates that the attention drawn from the visible region of the occluded object is essential to produce higher quality and semantically correct images. We performed various ablation studies to measure the effectiveness of each component of the loss function, and the results show that the perceptual loss and patch loss produce images with better quality and more semantic detail. The reconstructed images after adding the style loss contain more texture details.

In summary, our contributions are as follows: 1) We propose a weighted mask, which assigns more weight to the visible region of an occluded object and less attention to the rest of its containing image, and ignores the invisible pixels. 2) We train our model, which utilizes the weighted mask, to regenerate a partially visible instance through gated convolution and contextual attention in a self-supervised approach. We optimize the model with a combination of loss functions to produce a higher quality and more semantically correct output. 3) Our model can produce images with more semantic and texture details compared to the baseline models, specifically when the occluded object has a uniform texture. 

\section{Related Work}
\subsection{Amodal Content Completion}
There are several works that have focused on amodal completion. Ehsani et al. \cite{ehsani2018segan} propose SegGAN to segment the invisible area of objects by training the model on a synthetic dataset with photo-realistic images, then generate the appearance of the missing region using a conditional generative adversarial network (cGAN) \cite{mirza2014conditional}. To alleviate the need for amodal annotated training data, a self-supervised partial completion approach is implemented to train PCNets \cite{zhan2020self}. By randomly placing an arbitrary object on another object, the modal mask of both objects is obtained, which are then fed alternately to the model to complete the amodal mask of the occluded instance. The modal mask of the target object and the mask of the occluded area become the input to the partial content completion network, which learns to fill the RGB content of the missing patch. A more recent work \cite{ozguroglu2024pix2gestalt} performs amodal segmentation, recognition and 3D regeneration of occluded instances by fine-tuning pre-trained large-scale conditional diffusion models on a synthetically generated dataset with various occlusions.

\subsection{Gated Inpainting}
There is a large body of research on image inpainting in the literature that can be classified into several taxonomies based on the applied strategy \cite{zhang2023image} \cite{xiang2023deep}. One such category is convolution-aware inpainting, which defines approaches that alter the convolution filters to adapt with irregular holes in the input image. In vanilla convolutions, the same filter is applied across spatial positions. However, for images with holes, this results in poor reconstructed results with visual artifacts. The pixels in the input image are valid in the area outside the masked region, while the pixels in the masked area are invalid and should not be considered in the convolution operation. 

To impede the effect of invalid pixels on the repaired image, partial convolution is proposed by Liu et al. \cite{liu2018image} which implements a masking and a renormalization technique to apply convolution only on valid pixels of the image. Each traditional convolution layer is replaced by the partial convolutional layer such that the partial convolution operation is performed on the valid pixels indicated by the mask. After each convolution, the mask is updated in a way that if there are valid pixels in the current window, the current position becomes 1. Instead of the hard ruling for updating the mask, gated convolution \cite{yu2019free} learns mask automatically from the data through dynamic feature selection mechanism for each channel and at each spatial position. A light weight design of gated convolutions is adopted in \cite{yi2020contextual} that improves efficiency by decreasing the number of parameters required to automatically learn the mask. A bidirectional approach consisting of a forward attention map module and a reverse attention map module is implemented by the authors of \cite{xie2019image}. Whereas in the forward attention map module, the mask of the missing areas is used and updated in the encoding stage, the reverse attention map module utilizes the mask of the visible patches and updates it in the decoding stage. The masks specify the valid pixels in both the encoder and the decoder. On the other hand, region-wise convolutions are employed by Ma et al. \cite{ma2019coarse} in the decoder of their proposed model. In the decoding phase, two sets of convolutions are applied to reconstruct the content for the known and missing areas.

\section{Amodal Content Completion Model}
The conventional convolution operations that extract features from the pixels of a complete image are not applicable for images with holes. Since the input image contains both regions with valid pixels outside the holes and invalid pixels in masked areas, either hard-gating mechanism such as partial convolution \cite{liu2018image} or a soft gating mechanism such as gated convolution \cite{yu2019free} is used to enforce the convolution to depend only on valid pixels. The gated convolution essentially concatenates the input image and the mask of the invisible region in it (hereafter called the mask) and performs convolution on them to learn a dynamic feature gating mechanism. On each pixel located at ($x$,$y$) of the input feature map $I$, the output map ($O$) is computed by applying gated convolution as:
\begin{align}
G_{x,y} &= \sum_{u=-k_{h}}^{k_{h}} \sum_{v=-k_{w}}^{k_{w}} W^{\dagger}_{  k_{h}+u,k_{w}+v} \cdot I_{x+u,y+v} \\
F_{x,y} &= \sum_{u=-k_{h}}^{k_{h}} \sum_{v=-k_{w}}^{k_{w}} W^{\ddagger}_{k_{h}+u,k_{w}+v} \cdot I_{x+u,y+v} \\
O_{x,y} &= \phi(F_{x,y}) \odot \sigma(G_{x,y})
\end{align}
where $k_{h}$ = $\frac{h-1}{2}$, $k_{w}$ = $\frac{w-1}{2}$ are specified by the kernel size ($h \times w$). $W^{\dagger}$, $W^{\ddagger}$ are two separate convolutional filters. While $\phi$ can be any activation function, $\sigma$ represents the sigmoid function.

However, in amodal content completion we focus on completing an individual object specified by the amodal mask. The pixels in the occluded region are considered invalid pixels and should not be considered; instead, only the valid pixels outside the occluded patch must be utilized in the convolution. Based on our intuition, objects are semantically more similar throughout their different parts than the surrounding objects; therefore, the visible region of the object is more essential than the rest of the image.

In order to enforce the model to concentrate on the global features of the image and the visible region of the object concurrently, we employ the gated convolutions and contextual attention \cite{yu2018generative} with an additional mask, we call this \textit{weighted mask}. 

Instead of only having 0s and 1s in the mask, we add and utilize the weighted mask by assigning 0 to invalid pixels, 1 to the modal mask of the object, and 0.5 to the rest of the image. Given two instances $X$ and $Y$, along with their modal masks $M_X$ and $M_Y$ respectively, we randomly position $Y$ on $X$ to obtain the mask $M_{X \cap Y}$. The pixels in the weighted mask ($M_{weighted}$) of the image $I$ are then defined as:

\[
    M_{weighted} = 
\begin{cases}
    0,& \text{if } M_{X \cap Y} = 1\\
    1, & \text{if } M_{X \setminus Y} = 1\\
    0.5 & \text{otherwise}
\end{cases}
\]

where $M_{X \setminus Y}$ indicates the visible region of the instance $X$.

\begin{figure}[!h]
\centerline{\includegraphics[height=0.27\textwidth, width=0.4\textwidth]{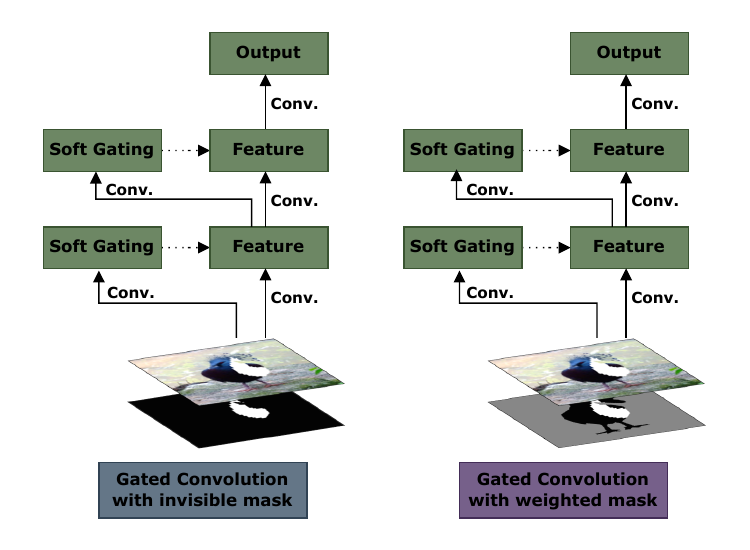}}
\caption{Gated convolution as used in DeepFill \cite{yu2019free} and ours with the weighted mask.}
\label{gatedConv}
\end{figure}

The incentive behind this assignment is that we want to extract the global features of the image and the visible region of the object concurrently, but with more weight assigned to the latter. 

The model consists of a coarse and a refinement network. Given the input image and the weighted mask, several layers of gated convolution are applied on them in the coarse network to produce an intermediate coarse image. The refinement network then takes the concatenation of the coarse image and the weighted mask and applies a series of gated convolutions and contextual attention layers to produce the completed image. The contextual attention layer learns where to borrow or copy feature information from visible background patches with similar features to produce the content of the missing patch. Finally, a spectral-normalized discriminator called SN-PatchGAN computes hinge loss on dense image patches (see Fig.~\ref{ourModel} for our model architecture).

\begin{figure*}[!h]
\centerline{\includegraphics[height=0.3\textwidth, width=0.75\textwidth]{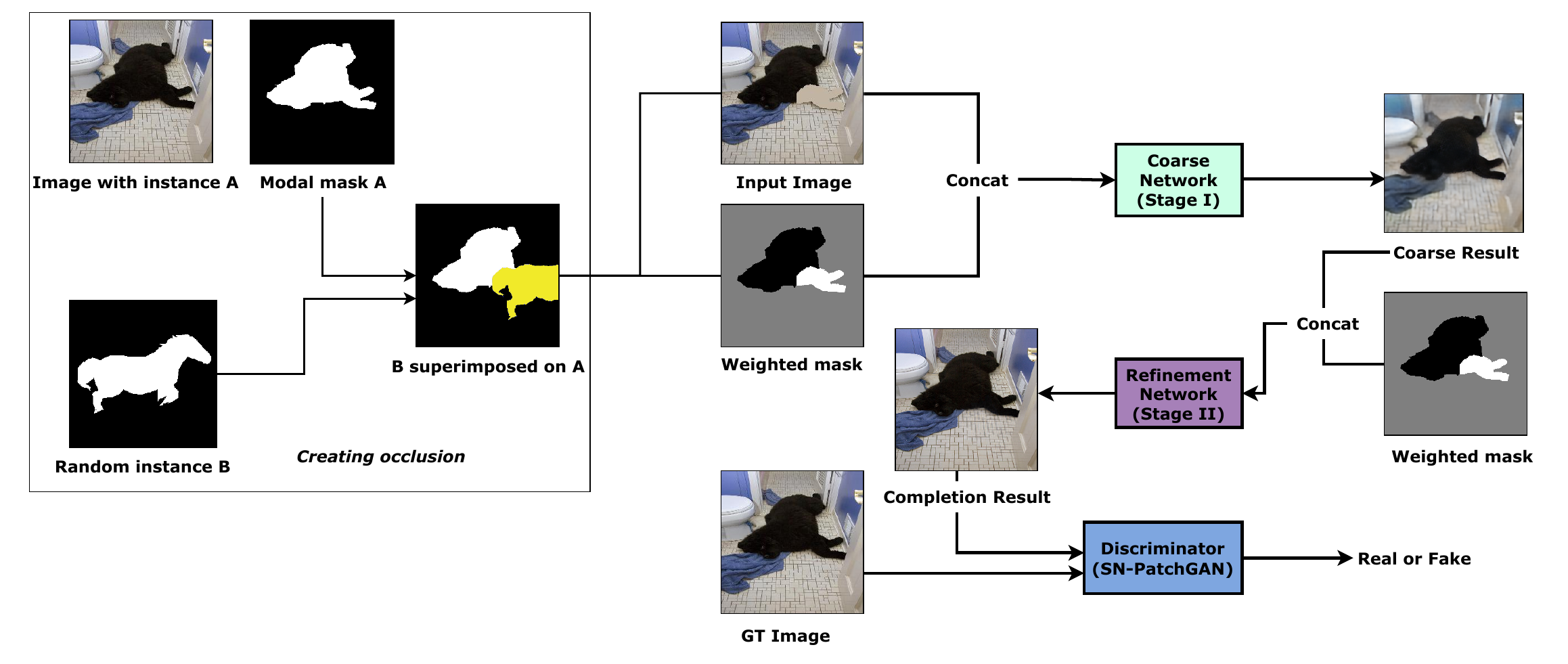}}
\caption{Our completion model with gated convolution and SN-PatchGAN as used in DeepFill \cite{yu2019free} with an additional weighted mask.}
\label{ourModel}
\end{figure*} 

\subsection{Training Data}
One of the biggest challenges of training a model for amodal content completion is the lack of labeled amodal training data. There is no dataset with RGB content for the missing region of the occluded instances. To overcome this challenge, we apply the self-supervised approach as described by Zhan et al. \cite{zhan2020self}. By randomly selecting an object from the dataset and positioning it on another object, we create occlusion. The amodal segmentation mask for training can then be obtained by merging the modal mask of the occluded object and the mask for the occluded region. The model is then used to complete the appearance of the occluded area indicated by the mask. 

To train our model, we use COCOA dataset. COCOA is a subset of COCO \cite{lin2014microsoft} dataset with 22163 instances from 2500 images in the training set, and 12753 instances from 1323 images in the validation set. We manually extract a subset from COCOA dataset with images that contain animal instances, we call it COCOA-animal-M. To increase the number of samples per instance, we perform augmentations using flipping, rotation, cropping, and shifting. After augmentation, the subset contains 3944 and 302 instances in the training and validation splits, respectively. We prepare another subset by filtering out instances with animal labels at the time of loading the dataset, we name it COCOA-animal-cls. With augmentations, this retains 7440 and 529 instances in the training split and the validation split, respectively.


\subsection{Loss Functions}\label{AA}

To train the generator, we employ five loss functions jointly: hinge loss \cite{lim2017geometric}, pixel-wise $\mathcal{L}$1 loss (also called reconstruction loss), perceptual loss \cite{liu2018image} for semantic similarity, patch loss (patch reconstruction loss) \cite{yang2017high}, and style loss \cite{xian2018texturegan}.

Hinge loss is used for training stability, and its objective function for generator (\textit{G}) is
\begin{linenomath}
\begin{equation}
\label{eq:hingeG}
\mathcal{L}_{G} = \--\ \mathbb{E}_{z \sim p_{z}(z)} [D(G(z))]
\end{equation}
\end{linenomath}
and for the discriminator (\textit{D}) is
\begin{linenomath}
\begin{equation}
\begin{split}
\label{eq:hingeD}
\mathcal{L}_{D} = \mathbb{E}_{x \sim p_{data}(x)} [min(0, \--\ 1 + D(x))] + \\ \mathbb{E}_{z \sim p_{z}(z)} [min(0, \--\ 1 \--\ D(G(z))]
\end{split}
\end{equation}
\end{linenomath}
where \textit{x} and \textit{z} are the ground-truth (GT) and the generated images, respectively. 

$\mathcal{L}$1 calculates the pixel-wise absolute difference between the GT image and the reconstructed images:

\begin{equation}
\label{eq:l1}
\mathcal{L}_{1} = \| x \--\ I_{out}^{1} \|_{1} + \| x \--\ I_{out}^{2} \|_{1}
\end{equation}
where \textit{x} is the GT image. $I_{out}^{1}$ and $I_{out}^{2}$ are the generated images from the coarse and refinement networks, respectively. 

On the other hand, the perceptual loss $\mathcal{L}_{perc}$ is computed by the absolute difference between the features of GT and the generated image:
\begin{linenomath}
\begin{equation}
\label{eq:percLoss}
\mathcal {L}_{perc} = \sum_{n=0}^{N - 1} \|\Psi_{n} (I_{out}) \--\ \Psi_{n} (I_{gt})\|1
\end{equation}
\end{linenomath}
$I_{out}$ and $I_{gt}$ represent the features of the output image and GT image, respectively. $\Psi_{n}$ is the activation function of the ${n}$th layer. The features are extracted from the fourth block of a pre-trained VGG-16 \cite{simonyan2014very} network (\textit{conv}4\textunderscore3).

Similarly, patch loss $\mathcal{L}_{patch}$ measures the absolute difference between the reconstructed patch and the corresponding original patch.
\begin{equation}
\label{eq:patchLoss}
\mathcal{L}_{patch} = \| M \odot (x \--\ I_{out}) \|_{1}
\end{equation}
where $M$ is the mask of the missing region. $\odot$ is the element-wise product operation. \textit{x} and $I_{out}$ are the GT and the generated image, respectively. 

Style loss $\mathcal{L}_{style}$ is applied to encourage the reconstruction of texture details. Similar to \cite{xian2018texturegan}, the Gram matrices (feature correlations) of features extracted from two layers of the VGG-19 \cite{simonyan2014very} network (\textit{relu}3\textunderscore2, and \textit{relu}4\textunderscore2) are calculated and used to define the style loss. The Gram matrix $\mathcal G_{ij}^{l} \in$ $\mathcal{R}^{{N_{l}}\times{N_{l}}}$ is defined as:
\begin{linenomath}
\begin{equation}
\label{eq:styleLoss}
\mathcal G_{ij}^{l} = \sum_{k} \mathcal{F}_{ik}^{l} \mathcal{F}_{jk}^{l}
\end{equation}
\end{linenomath}
where $N_{l}$ is the number of feature maps at network layer ${l}$, $\mathcal{F}_{jk}^{l}$ is the activation of the $i$th filter at position ${k}$ in layer ${l}$.

Taking the previous loss functions into account, the loss of our model to minimize is formulated as follows:
\begin{equation}
\label{eq:totalLoss}
\mathcal{L}_{total} = \lambda_{1} \mathcal{L}_{G} + \lambda_{2} \mathcal{L}_{perc} + \lambda_{3} \mathcal{L}_{patch} + \lambda_{4} \mathcal{L}_{style} + \lambda_{5} \mathcal{L}1
\end{equation}
where $\lambda_{1}$, $\lambda_{2}$, $\lambda_{3}$, $\lambda_{4}$, and $\lambda_{5}$ represent the weight factor of the loss terms. We set $\lambda_{1}$ = 1, $\lambda_{2}$ = 100, $\lambda_{3}$ = 10, $\lambda_{4}$ = 1, and $\lambda_{5}$ = 100.

\section{Results and Discussion}
We evaluate our method in amodal content completion on the validation set of COCOA, COCOA-animal-M, and COCOA-animal-cls datasets.  

We report the quantitative results in Table~\ref{table:results}. The baselines include PCNet-C \cite{zhan2020self} and DeepFill \cite{yu2019free}. Both baselines were adjusted when trained to obtain their respective best results. The completed images are evaluated in terms of mean $\mathcal{L}_{1}$ and $\mathcal{L}_{2}$ errors, PSNR, and SSIM metrics.

\begin{table}[!h]
\caption{Quantitative results of PCNet-C, DeepFill, and our model on the validation set of COCOA and its two subsets. Our model is evaluated with all five components of the loss function and then without each one of them (except for the hinge loss).}
\label{table:results}
\centering
    \begin{tabular}{ p{0.28\linewidth} p{0.15\linewidth} p{0.15\linewidth} p{0.11\linewidth} p{0.10\linewidth} } \hline  
     Description & $\mathcal{L}_{1}$ Error $\downarrow$ & $\mathcal{L}_{2}$ Error $\downarrow$ & PSNR $\uparrow$ & SSIM $\uparrow$ \\ 
     \hline
        \multicolumn{2}{l}{COCOA-animal-M}\\ \hline
        PCNet-C & 0.09541 & 0.01392 & 24.14104 & 0.80486 \\
        DeepFill & 0.05743 & 0.00928 & 24.95441 & 0.81857\\ 
        Ours w/o patch loss & 0.02617 & 0.00238 & 28.19811 & 0.93972 \\
        Ours w/o style loss & 0.02609 & 0.00245 & 28.22504 & 0.93904 \\ 
        Ours w/o perceptual loss & 0.05880 & 0.00880 & 22.50169 & 0.77525 \\
        Ours w/o $\mathcal{L}$1 loss & 0.03482 & 0.00338 & 26.91633 & 0.92547 \\
        Ours with four loss terms & \textbf{0.02521} & \textbf{0.0023} & \textbf{28.29431} & \textbf{0.94257} \\
        \hline \hline
     
        \multicolumn{2}{l}{COCOA-animal-cls}\\ \hline
        PCNet-C & 0.11482 & 0.01783 & 23.04228 & 0.73566 \\
        DeepFill & 0.02847 & 0.00359 & 29.37148 & 0.92870\\ 
        Ours w/o patch loss & 0.02155 & 0.00183 & 28.95409 & 0.94213 \\ 
        Ours w/o style loss & 0.02143 & 0.00185 & 29.00787 & 0.94233 \\ 
        Ours w/o perceptual loss & 0.04336 & 0.00564 & 24.38351 & 0.82619 \\ 
        Ours w/o $\mathcal{L}$1 loss & 0.02835 & 0.00257 & 28.00027 & 0.93453 \\ 
        Ours with four loss terms & \textbf{0.02007} & \textbf{0.00174} & \textbf{29.41134} & \textbf{0.94616} \\ \hline \hline
       
        \multicolumn{2}{l}{Original COCOA}\\ \hline
        PCNet-C & 0.03539 & 0.00296 & 30.64838 & 0.91799 \\
        DeepFill & \textbf{0.01367} & 0.00141 & \textbf{31.836} & \textbf{0.96007}\\
        Ours & 0.01441 & \textbf{0.00113} & 31.62578 & 0.9549 \\
        \hline \hline
    \end{tabular}
    
\end{table}

The results indicate that our model leads to lower mean error and higher-quality reconstruction on both COCOA-animal-M and COCOA-animal-cls datasets. It can be seen from Fig.~\ref{fig:outputs} that our model can generate higher semantic and texture details when completing the partially visible
objects compared to the baseline models. This illustrates that the
attention drawn from the visible region of the occluded object
is essential in producing higher quality and semantically correct
images. 
\begin{figure*}[!h]\captionsetup[subfigure]{font=tiny}
\centering
		\begin{subfigure}{0.12\textwidth}
                \includegraphics[width=\textwidth, frame]{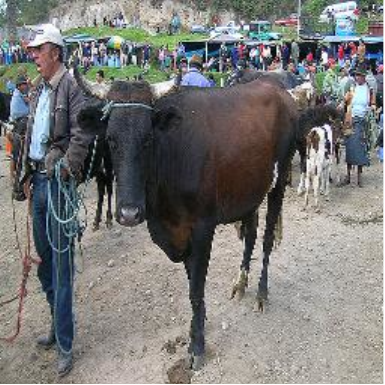}                
                 \includegraphics[width=\textwidth, frame]{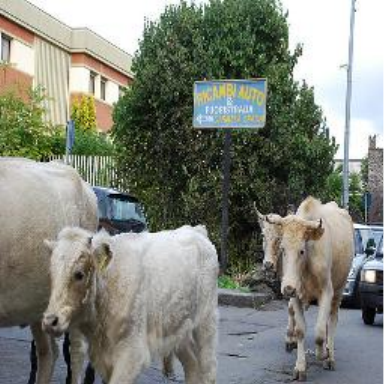}
                 \includegraphics[width=\textwidth, frame]{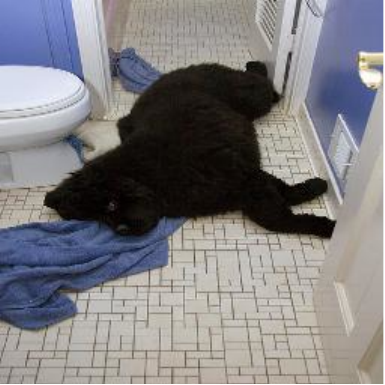}
                 \includegraphics[width=\textwidth, frame]{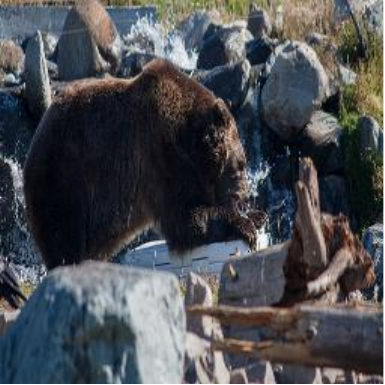}
                 \includegraphics[width=\textwidth, frame]{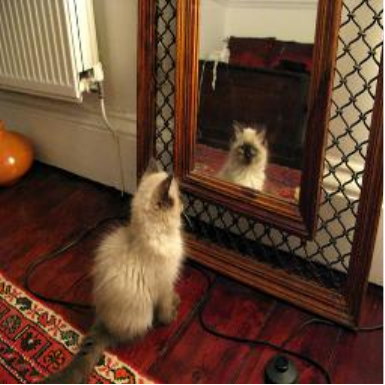}
                 \includegraphics[width=\textwidth, frame]{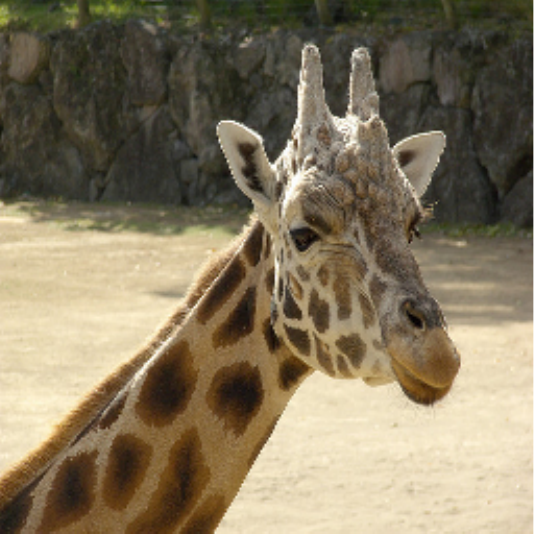}
			\caption{GT}
		\end{subfigure}
		\begin{subfigure}{0.12\textwidth}
                \includegraphics[width=\textwidth, frame]{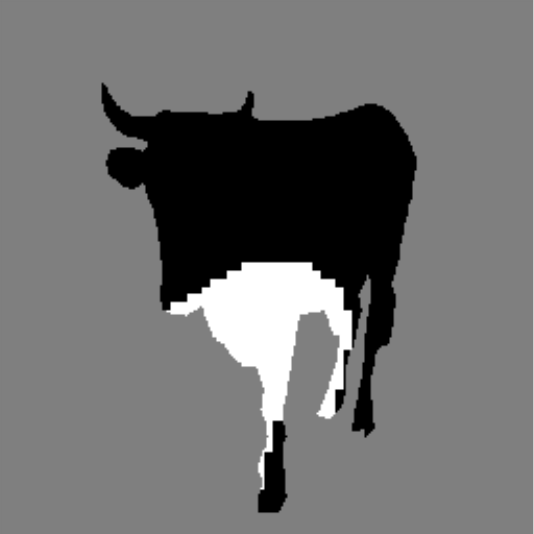}
                \includegraphics[width=\textwidth, frame]{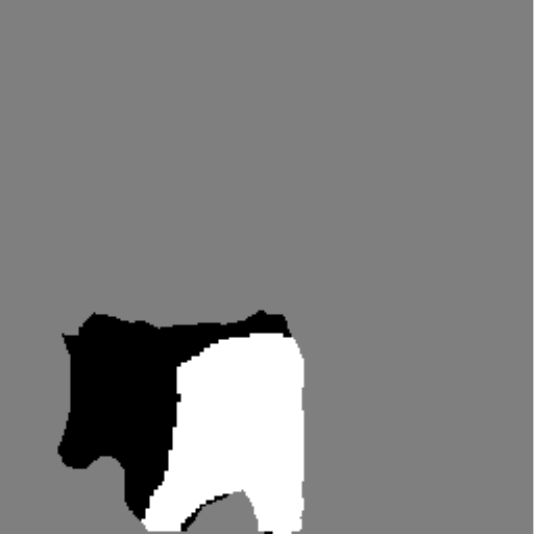}
                \includegraphics[width=\textwidth, frame]{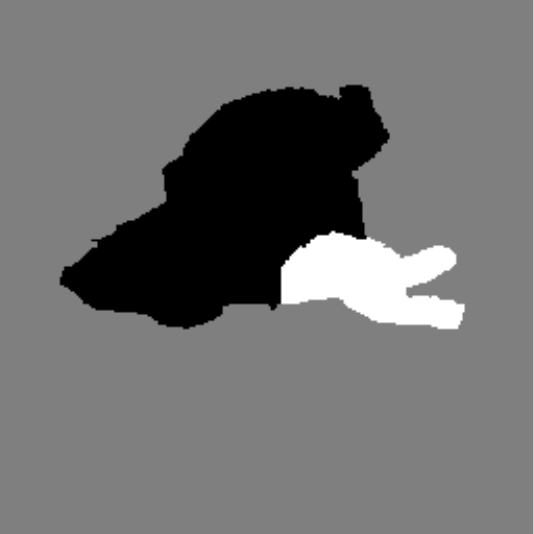}
                \includegraphics[width=\textwidth, frame]{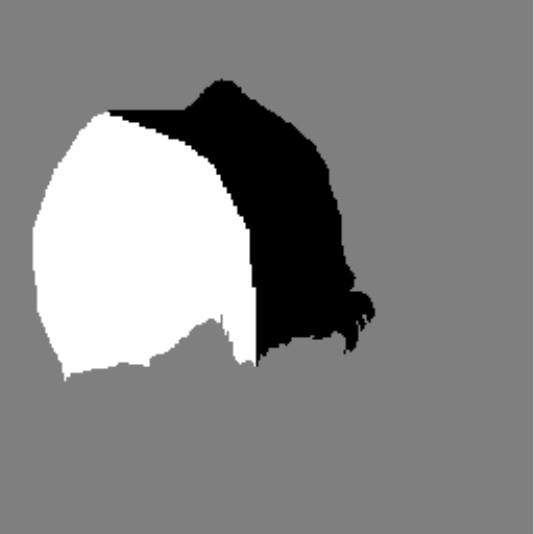}
                \includegraphics[width=\textwidth, frame]{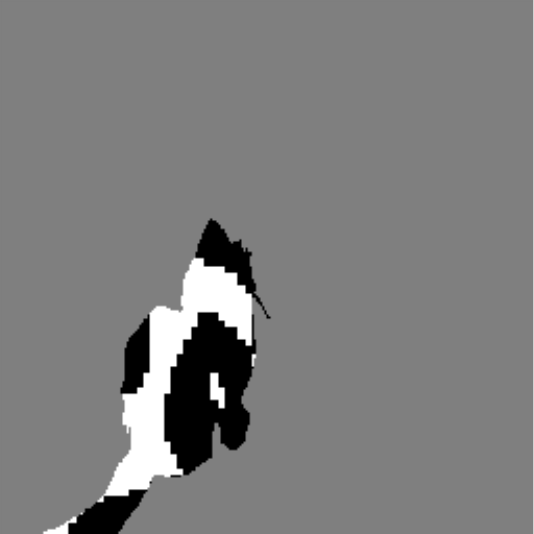}
                \includegraphics[width=\textwidth, frame]{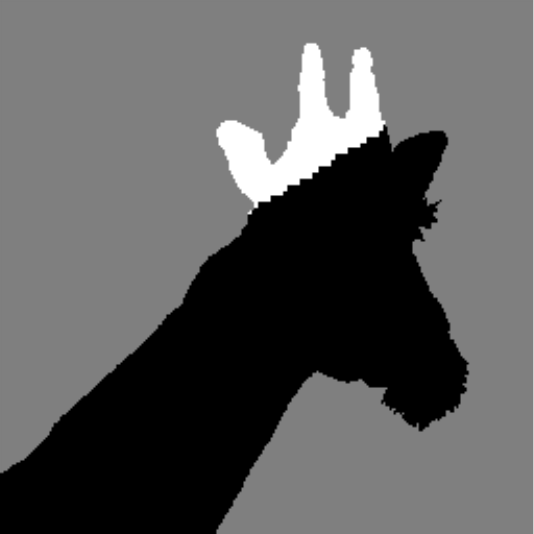}
			\caption{Weighted mask}
		\end{subfigure}
            \begin{subfigure}{0.12\textwidth}
                \includegraphics[width=\textwidth, frame]{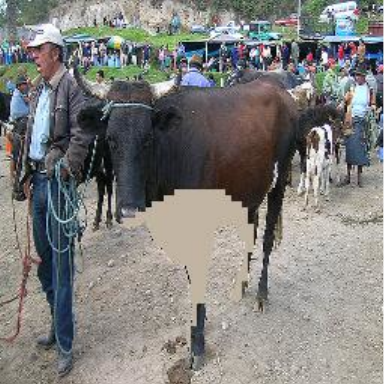}
                \includegraphics[width=\textwidth, frame]{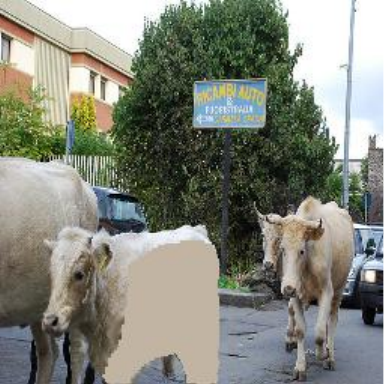}
                \includegraphics[width=\textwidth, frame]{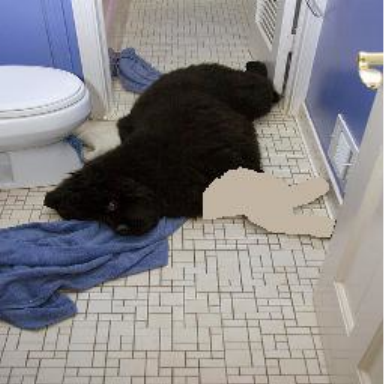}
                \includegraphics[width=\textwidth, frame]{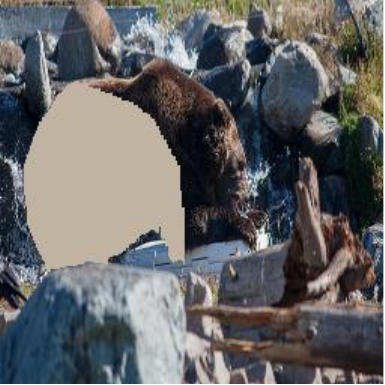}
                \includegraphics[width=\textwidth, frame]{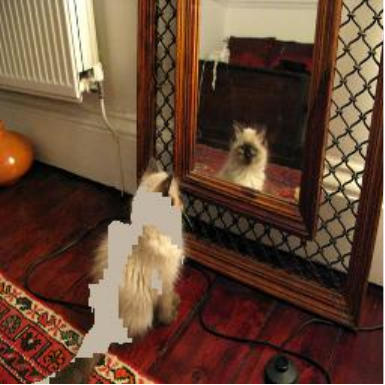}
                \includegraphics[width=\textwidth, frame]{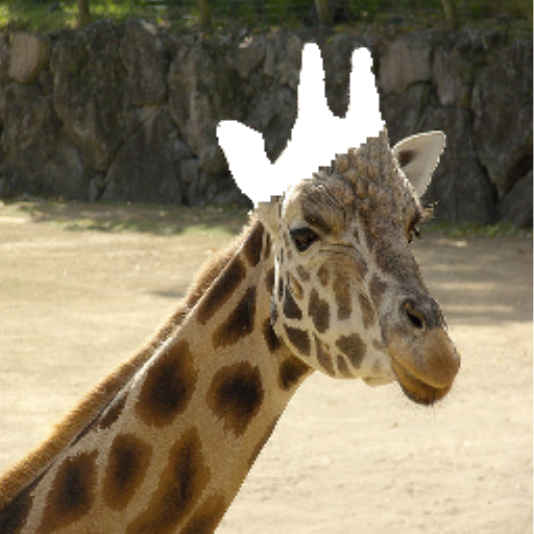}
			\caption{Masked image}
		\end{subfigure}
            \begin{subfigure}{0.12\textwidth}
                \includegraphics[width=\textwidth, frame]{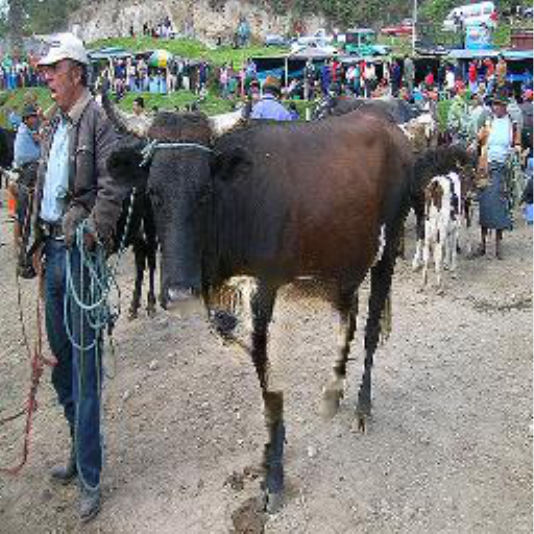}
                \includegraphics[width=\textwidth, frame]{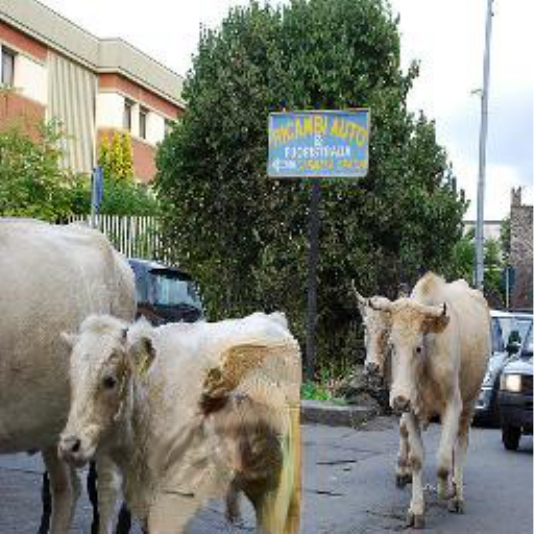}
                \includegraphics[width=\textwidth, frame]{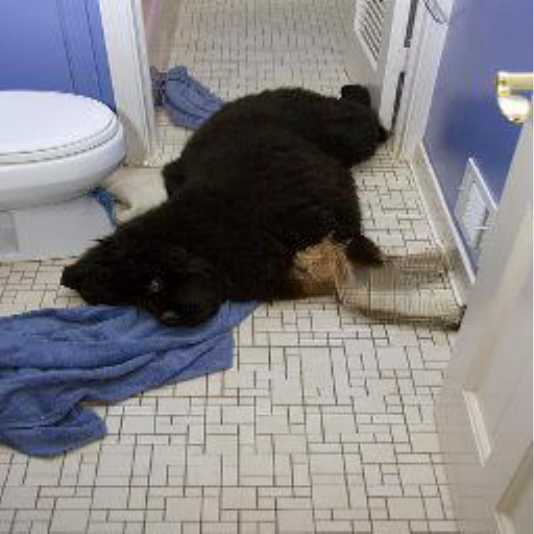}
                \includegraphics[width=\textwidth, frame]{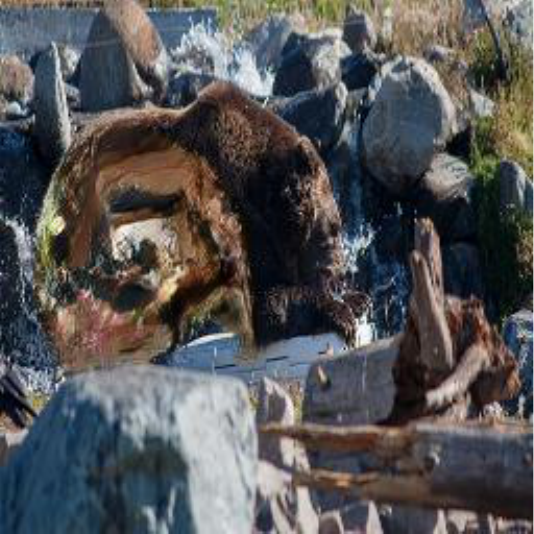}
                \includegraphics[width=\textwidth, frame]{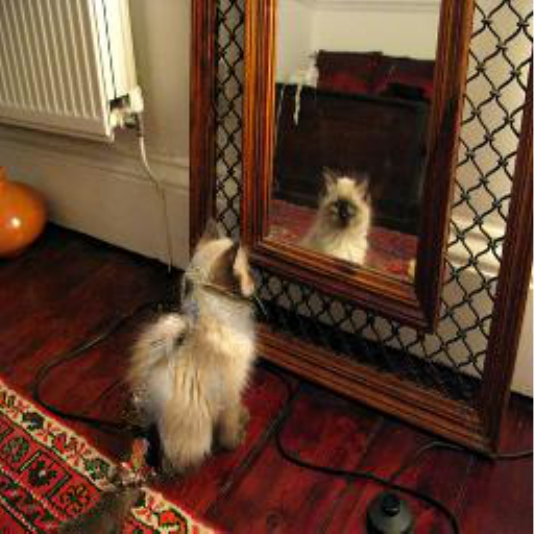}
                \includegraphics[width=\textwidth, frame]{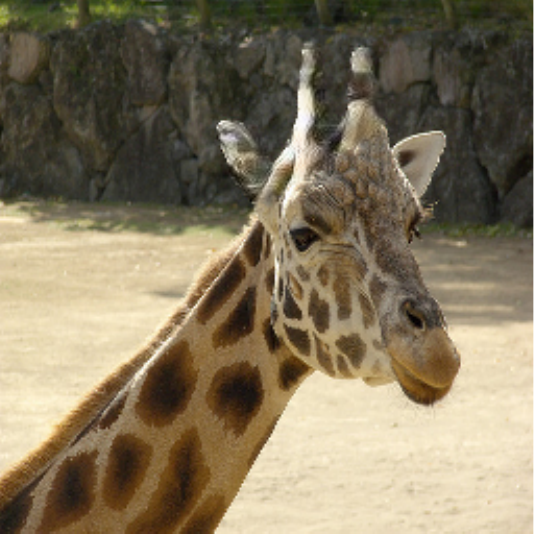}
			\caption{DeepFill (pre-trained)}
		\end{subfigure}
            \begin{subfigure}{0.12\textwidth}
                \includegraphics[width=\textwidth, frame]{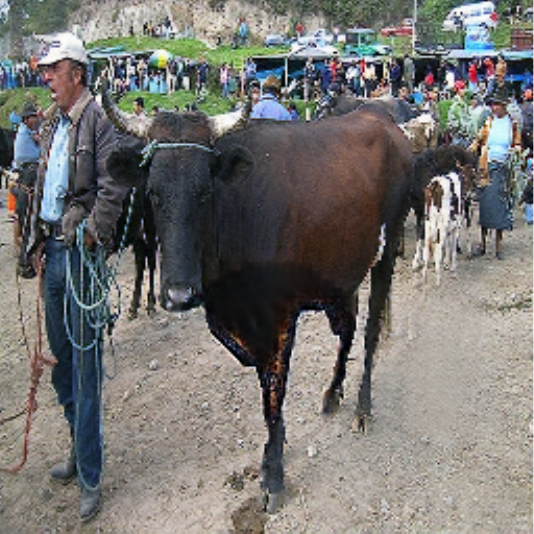}
                \includegraphics[width=\textwidth, frame]{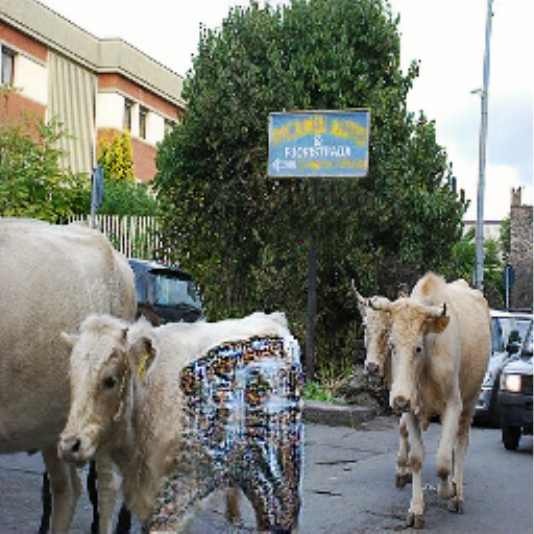}
                \includegraphics[width=\textwidth, frame]{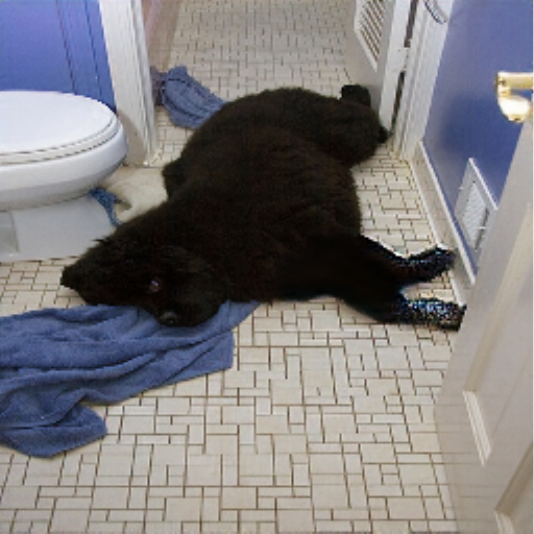}
                \includegraphics[width=\textwidth, frame]{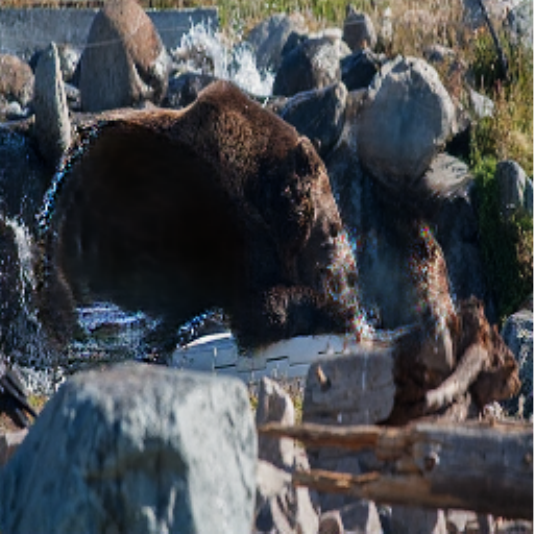}
                \includegraphics[width=\textwidth, frame]{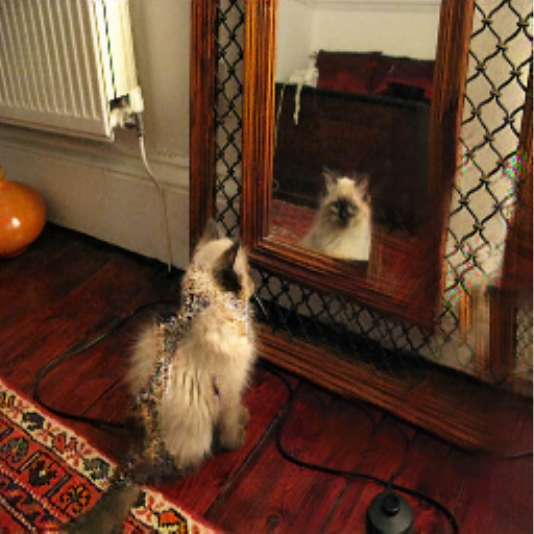}
                \includegraphics[width=\textwidth, frame]{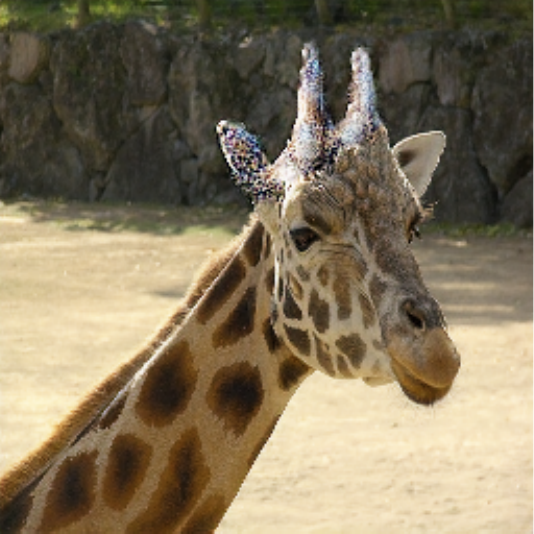}
			\caption{DeepFill (re-trained)}
		\end{subfigure}
		\begin{subfigure}{0.12\textwidth}
               \includegraphics[width=\textwidth, frame]{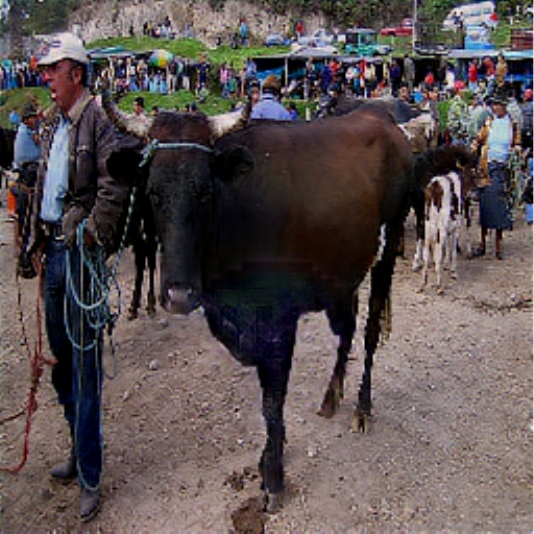}
                \includegraphics[width=\textwidth, frame]{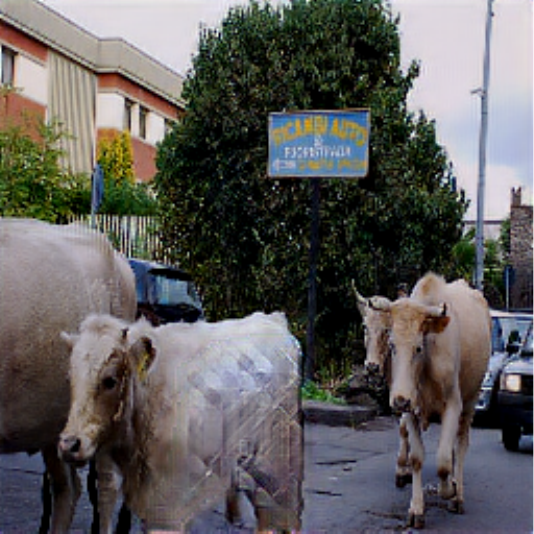}
                \includegraphics[width=\textwidth, frame]{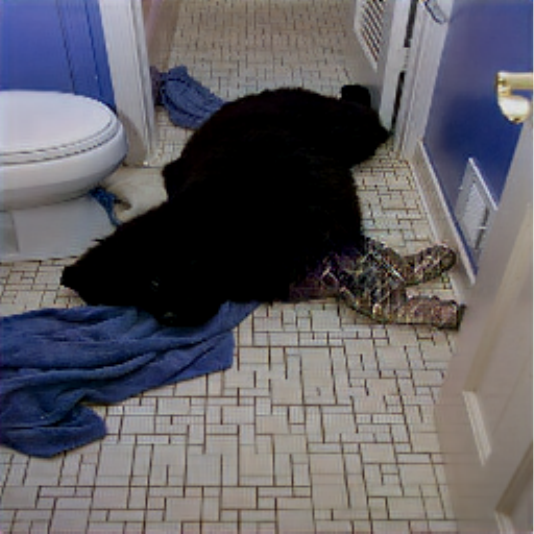}
                 \includegraphics[width=\textwidth, frame]{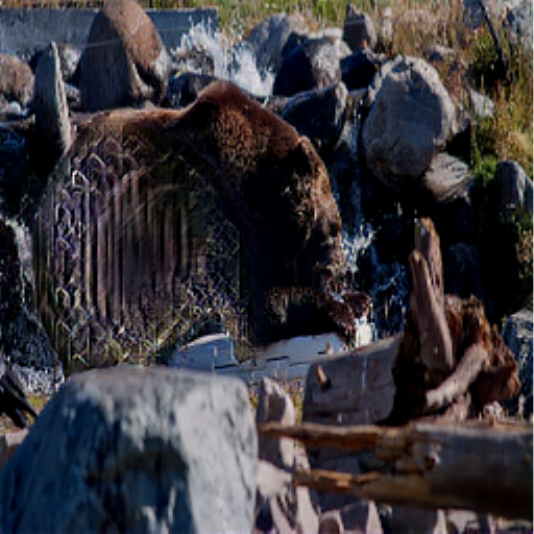}
                \includegraphics[width=\textwidth, frame]{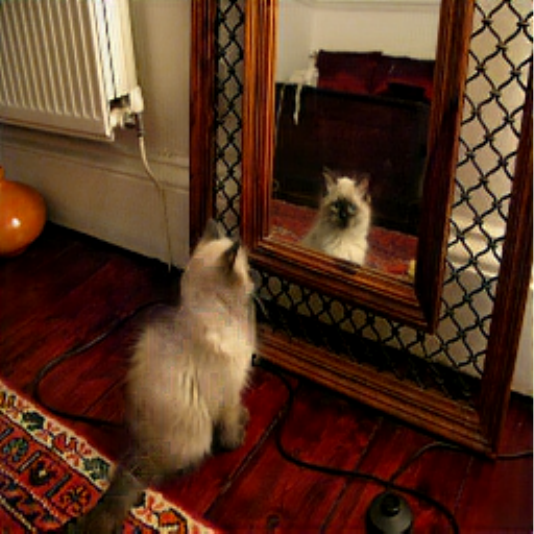}
                \includegraphics[width=\textwidth, frame]{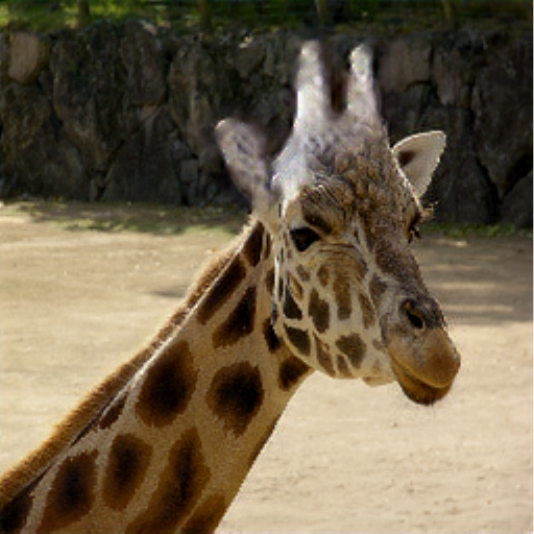}
			\caption{PCNet-C}
		\end{subfigure}
		\begin{subfigure}{0.12\textwidth}
                \includegraphics[width=\textwidth, frame]{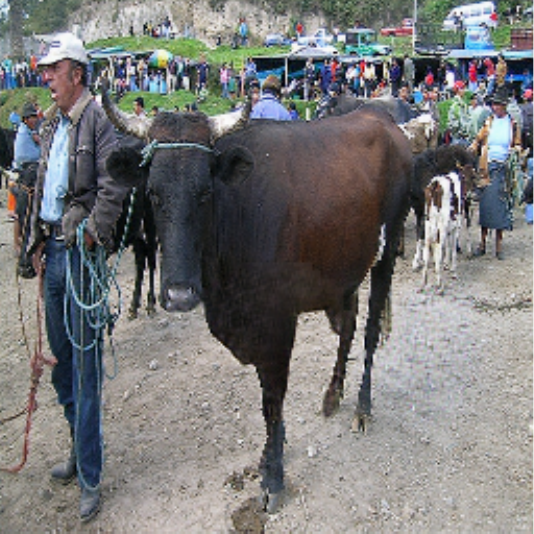}
                \includegraphics[width=\textwidth, frame]{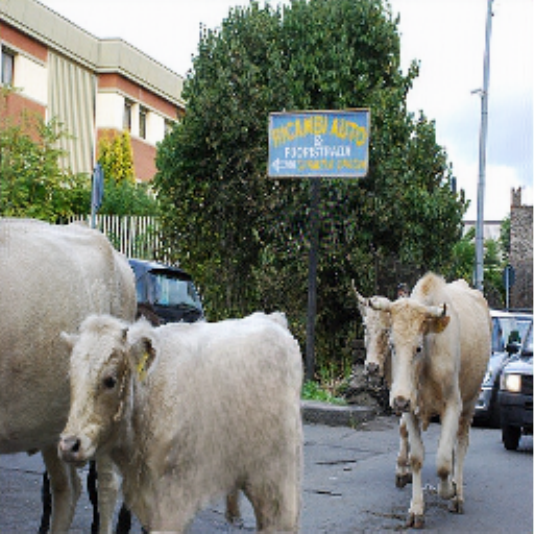}
                \includegraphics[width=\textwidth, frame]{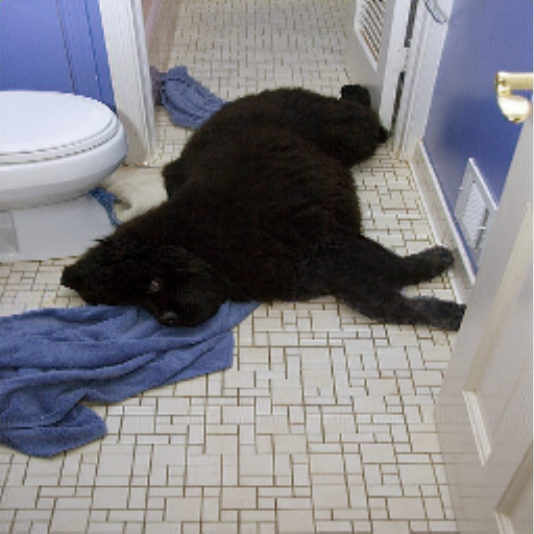}
                \includegraphics[width=\textwidth, frame]{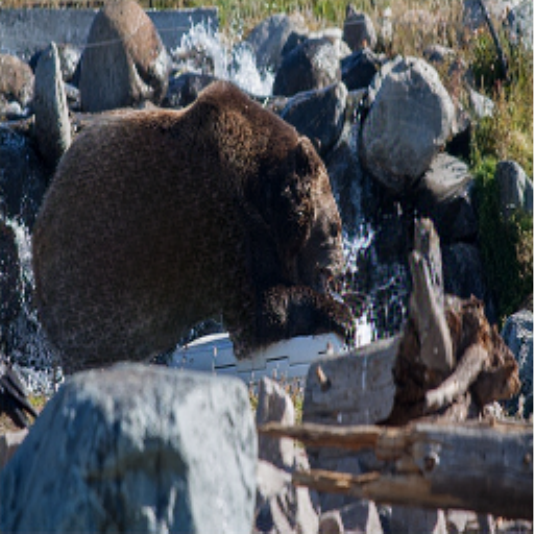}
                \includegraphics[width=\textwidth, frame]{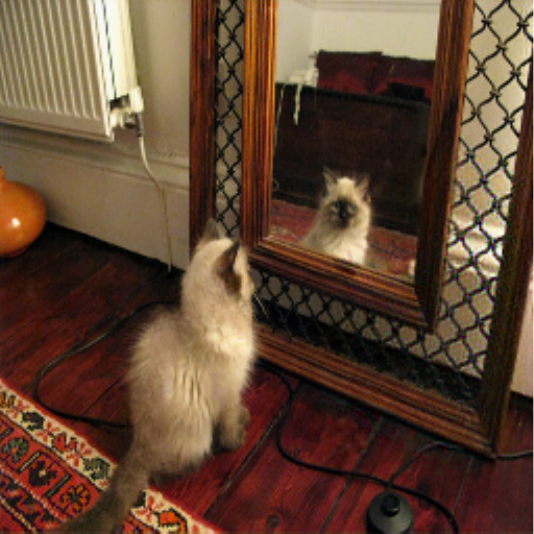}
                \includegraphics[width=\textwidth, frame]{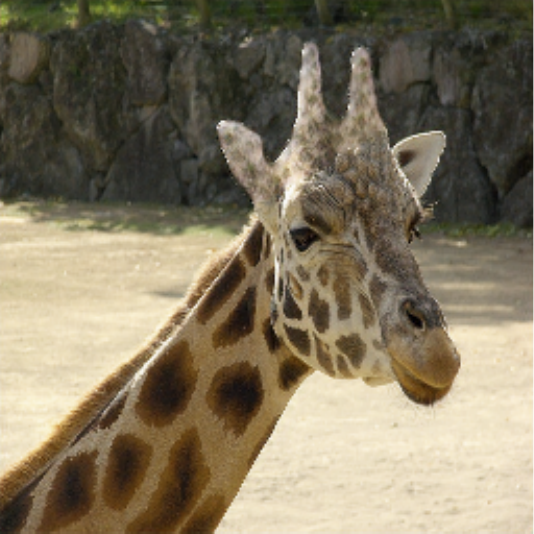}
			\caption{Ours}
		\end{subfigure}
				\caption{Qualitative comparison of completed images from COCOA-animal-M validation set. The weighted mask is used in our model, DeepFill and PCNet-C only utilize the mask of the missing region (indicated by the white pixels). (d) and (e) show results of the published version of DeepFill and the re-trained version on the used datasets, respectively.}
		\label{fig:outputs}
	\end{figure*}

We conduct several ablation studies to evaluate the effectiveness of each term of the loss function. Although according to Yu et al. \cite{yu2019free} perceptual loss is unnecessary since patch-level information is already encoded in SN-PatchGAN \cite{yu2019free}, our experiments show that this loss term is essential in decreasing the mean error rate and enhancing the quality of the output images (see Table~\ref{table:results} and Fig.~\ref{fig:ablation}). Similarly, patch loss and style loss produce images with more semantic and texture details. 

\begin{figure*}[!h]\captionsetup[subfigure]{font=tiny}
\centering
		\begin{subfigure}{0.12\textwidth}               
                 \includegraphics[width=\textwidth, frame]{Images/2_1_35_gt.pdf}
                 \includegraphics[width=\textwidth, frame]{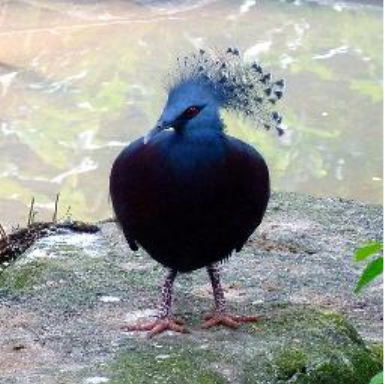}
                 \includegraphics[width=\textwidth, frame]{Images/2_1_8_gt.pdf}
			\caption{GT}
		\end{subfigure}
		\begin{subfigure}{0.12\textwidth}
                \includegraphics[width=\textwidth, frame]{Images/2_1_35_amodal.pdf}
                \includegraphics[width=\textwidth, frame]{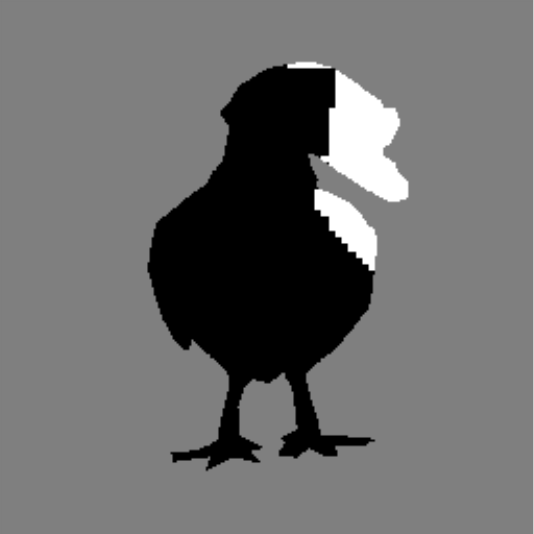}
                \includegraphics[width=\textwidth, frame]{Images/2_1_8_amodal.pdf}
			\caption{weighted mask}
		\end{subfigure}
            \begin{subfigure}{0.12\textwidth}
                \includegraphics[width=\textwidth, frame]{Images/2_1_35_masked.pdf}
                \includegraphics[width=\textwidth, frame]{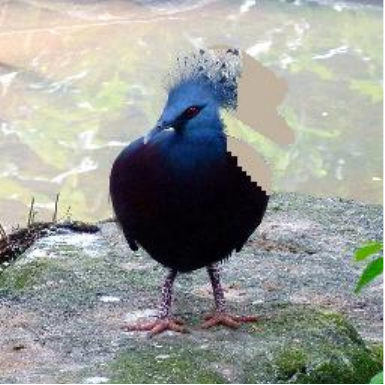}
                \includegraphics[width=\textwidth, frame]{Images/2_1_8_masked.pdf}
			\caption{Masked image}
		\end{subfigure}
            \begin{subfigure}{0.12\textwidth}
                \includegraphics[width=\textwidth, frame]{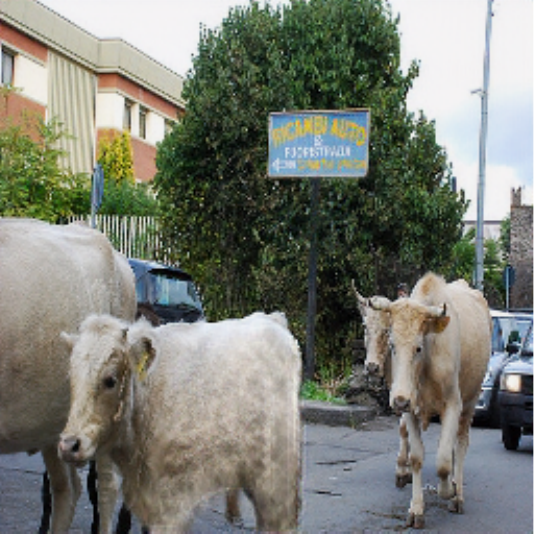}
                \includegraphics[width=\textwidth, frame]{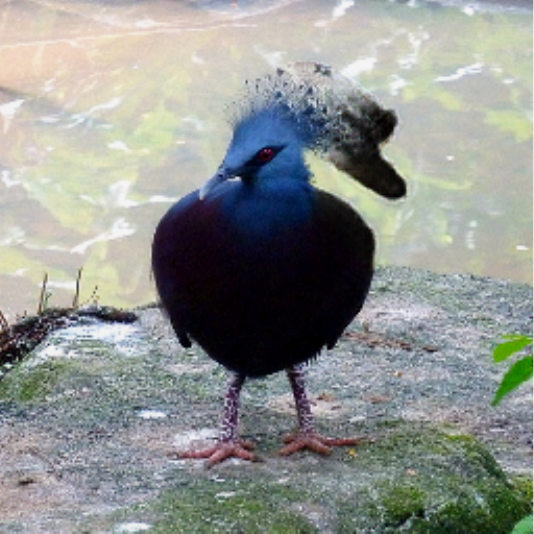}
                \includegraphics[width=\textwidth, frame]{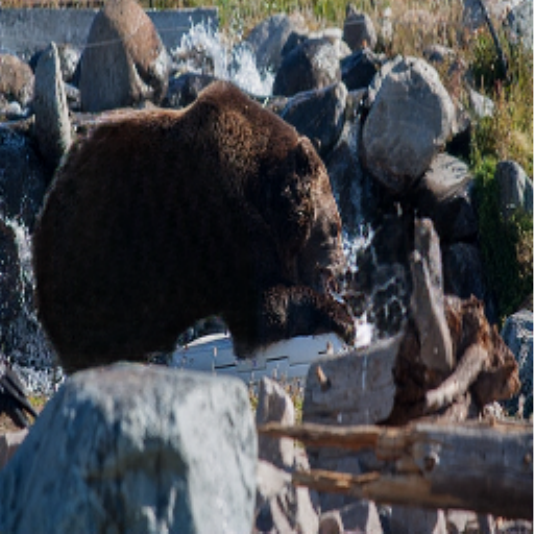}
			\caption{Ours w/o $\mathcal{L}$1 loss}
		\end{subfigure}
            \begin{subfigure}{0.12\textwidth}
                \includegraphics[width=\textwidth, frame]{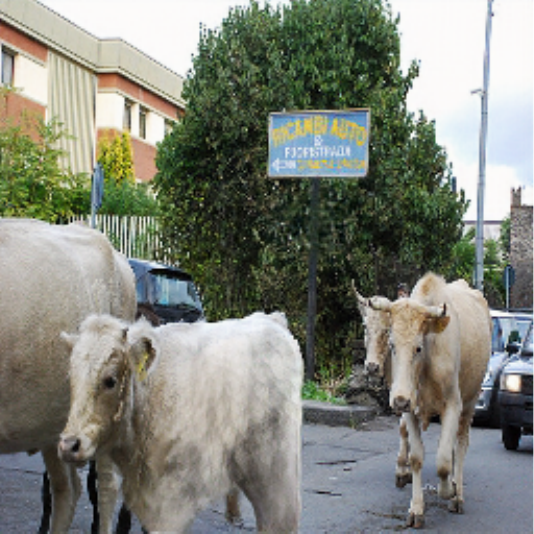}
                \includegraphics[width=\textwidth, frame]{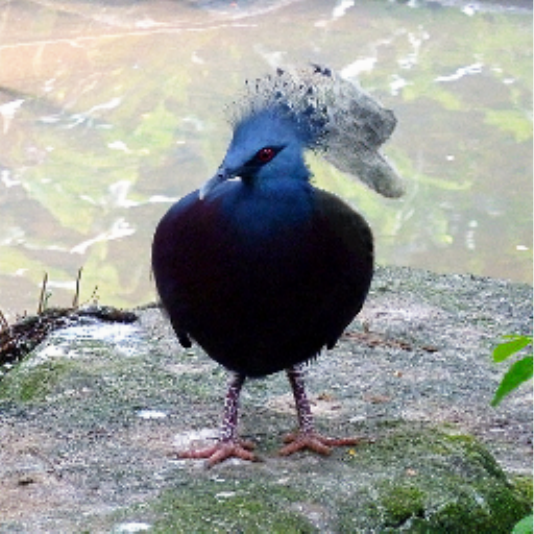}
                \includegraphics[width=\textwidth, frame]{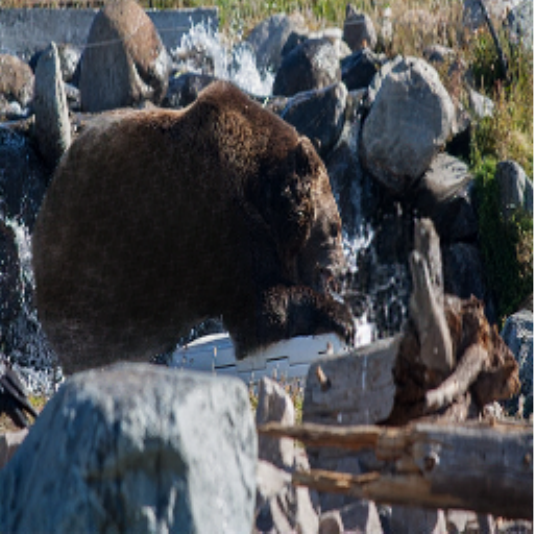}
			\caption{Ours w/o patch loss}
		\end{subfigure}
		\begin{subfigure}{0.12\textwidth}
                \includegraphics[width=\textwidth, frame]{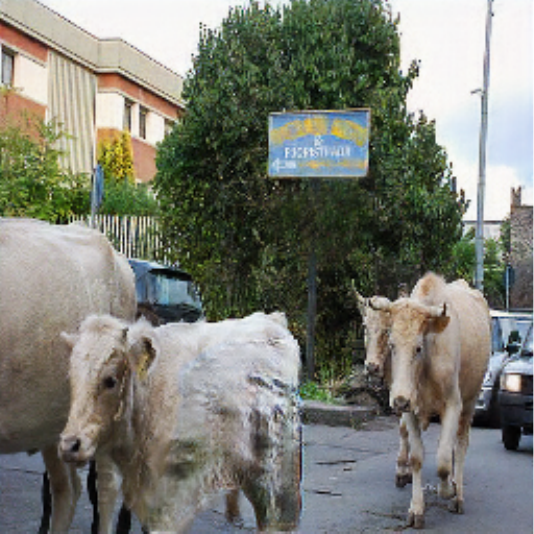}
                \includegraphics[width=\textwidth, frame]{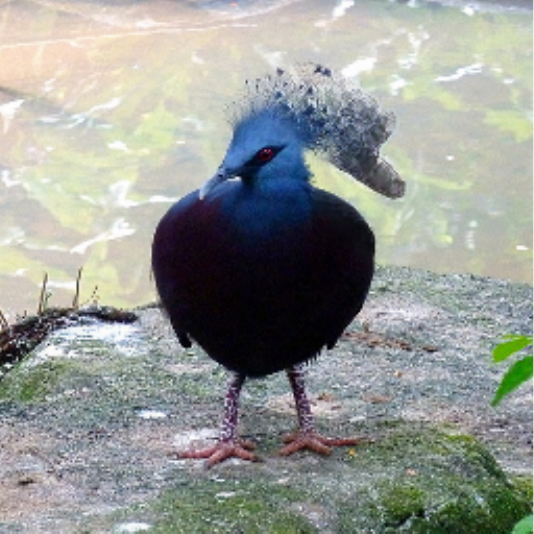}
                 \includegraphics[width=\textwidth, frame]{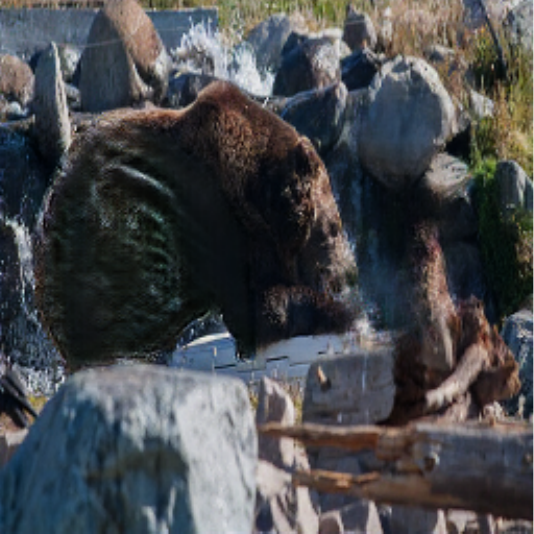}
			\caption{Ours w/o perceptual loss}
		\end{subfigure}
		\begin{subfigure}{0.12\textwidth}
                \includegraphics[width=\textwidth, frame]{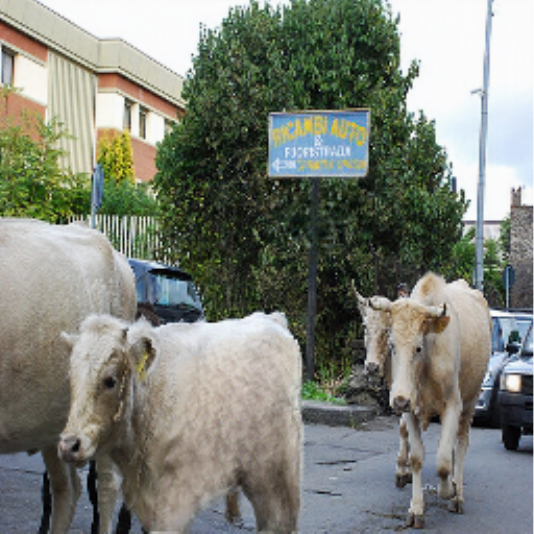}
                \includegraphics[width=\textwidth, frame]{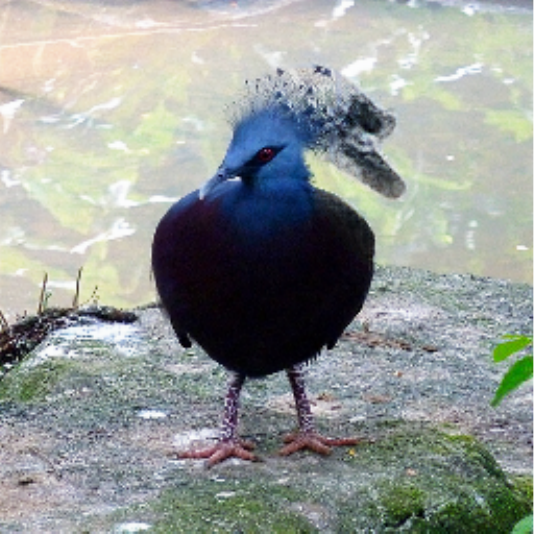}
                \includegraphics[width=\textwidth, frame]{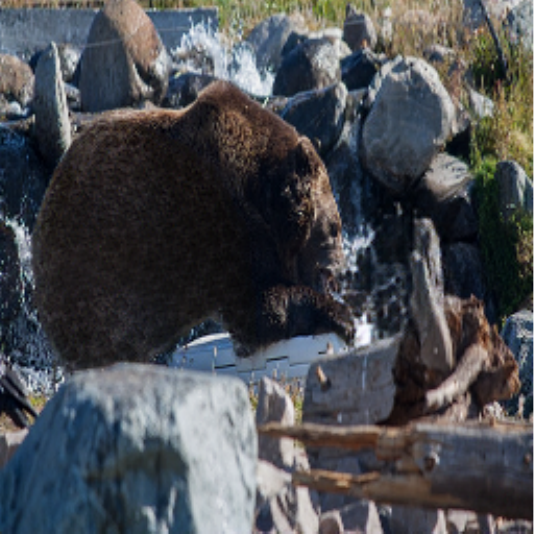}
			\caption{Ours w/o style loss}
		\end{subfigure}
				\caption{Ablation study of our model without each loss component.}
		\label{fig:ablation}
	\end{figure*}

Furthermore, our model performs slightly below DeepFill on the original COCOA dataset which contains instances of various categories other than animal category, and these include very small instances too. In the case of small objects, there may not be sufficient information in the visible region to aid the completion process. Similarly, in cases that include objects with disparate parts. such as humans, our approach does not produce satisfactory output.

\section{Conclusions}
We have proposed a model that utilizes gated convolution and a weighted mask to perform amodal content completion in a self-supervised manner. The weighted mask assigns more attention towards the visible region of an occluded instance than the other regions of the image, while filtering out the invisible patch. The model is trained with hinge loss, perceptual loss, pixel-wise $\mathcal{L}$1 loss, style loss, and patch loss. The results demonstrate that our model can produce a semantically correct output and retain more texture detail when completing a partially visible object than the baseline models. However, if the occluded instance has disparate parts, the model does not produce satisfactory results. 

 \section*{Acknowledgment}
We would like to thank the ``Doctoral School of Applied Informatics and Applied Mathematics'' and the ``High Performance Computing Research Group'' of \'{O}buda University for their valuable support. We also thank NVIDIA Corporation for
providing graphics hardware for the experiments. On behalf of the ``Development of Machine Learning Aided Metaheuristics'' project, we are grateful for the possibility to use HUN-REN Cloud \cite{heder2022past} which helped us achieve the results published in this paper.

\bibliographystyle{ieeetr}
\bibliography{conference_101719}

\begin{thebibliography}{10}

\bibitem{saleh2021occlusion}
K.~Saleh, S.~Sz{\'e}n{\'a}si, and Z.~V{\'a}mossy, ``Occlusion handling in generic object detection: A review,'' in {\em 2021 IEEE 19th World Symposium on Applied Machine Intelligence and Informatics (SAMI)}, pp.~000477--000484, IEEE, 2021.

\bibitem{ao2023image}
J.~Ao, Q.~Ke, and K.~A. Ehinger, ``Image amodal completion: A survey,'' {\em Computer Vision and Image Understanding}, p.~103661, 2023.

\bibitem{zhu2017semantic}
Y.~Zhu, Y.~Tian, D.~Metaxas, and P.~Doll{\'a}r, ``Semantic amodal segmentation,'' in {\em Proceedings of the IEEE conference on computer vision and pattern recognition}, pp.~1464--1472, 2017.

\bibitem{bs3}
L.~Qi, L.~Jiang, S.~Liu, X.~Shen, and J.~Jia, ``Amodal instance segmentation with kins dataset,'' in {\em Proceedings of the IEEE/CVF Conference on Computer Vision and Pattern Recognition}, pp.~3014--3023, 2019.

\bibitem{bs7}
Y.~Xiao, Y.~Xu, Z.~Zhong, W.~Luo, J.~Li, and S.~Gao, ``Amodal segmentation based on visible region segmentation and shape prior,'' in {\em Proceedings of the AAAI Conference on Artificial Intelligence}, vol.~35, pp.~2995--3003, 2021.

\bibitem{bs13}
M.~Tran, K.~Vo, K.~Yamazaki, A.~Fernandes, M.~Kidd, and N.~Le, ``Aisformer: Amodal instance segmentation with transformer,'' {\em arXiv preprint arXiv:2210.06323}, 2022.

\bibitem{bs16}
J.~Gao, X.~Qian, Y.~Wang, T.~Xiao, T.~He, Z.~Zhang, and Y.~Fu, ``Coarse-to-fine amodal segmentation with shape prior,'' in {\em Proceedings of the IEEE/CVF International Conference on Computer Vision}, pp.~1262--1271, 2023.

\bibitem{bs21}
G.~Zhan, C.~Zheng, W.~Xie, and A.~Zisserman, ``Amodal ground truth and completion in the wild,'' {\em arXiv preprint arXiv:2312.17247}, 2023.

\bibitem{bs22}
J.~Ao, Q.~Ke, and K.~A. Ehinger, ``Amodal intra-class instance segmentation: New dataset and benchmark,'' {\em arXiv preprint arXiv:2303.06596}, 2023.

\bibitem{ozguroglu2024pix2gestalt}
E.~Ozguroglu, R.~Liu, D.~Sur{\'\i}s, D.~Chen, A.~Dave, P.~Tokmakov, and C.~Vondrick, ``pix2gestalt: Amodal segmentation by synthesizing wholes,'' {\em arXiv preprint arXiv:2401.14398}, 2024.

\bibitem{zhan2020self}
X.~Zhan, X.~Pan, B.~Dai, Z.~Liu, D.~Lin, and C.~C. Loy, ``Self-supervised scene de-occlusion,'' in {\em Proceedings of the IEEE/CVF Conference on Computer Vision and Pattern Recognition}, pp.~3784--3792, 2020.

\bibitem{ehsani2018segan}
K.~Ehsani, R.~Mottaghi, and A.~Farhadi, ``Segan: Segmenting and generating the invisible,'' in {\em Proceedings of the IEEE conference on computer vision and pattern recognition}, pp.~6144--6153, 2018.

\bibitem{mirza2014conditional}
M.~Mirza and S.~Osindero, ``Conditional generative adversarial nets,'' {\em arXiv preprint arXiv:1411.1784}, 2014.

\bibitem{zhang2023image}
X.~Zhang, D.~Zhai, T.~Li, Y.~Zhou, and Y.~Lin, ``Image inpainting based on deep learning: A review,'' {\em Information Fusion}, vol.~90, pp.~74--94, 2023.

\bibitem{xiang2023deep}
H.~Xiang, Q.~Zou, M.~A. Nawaz, X.~Huang, F.~Zhang, and H.~Yu, ``Deep learning for image inpainting: A survey,'' {\em Pattern Recognition}, vol.~134, p.~109046, 2023.

\bibitem{liu2018image}
G.~Liu, F.~A. Reda, K.~J. Shih, T.-C. Wang, A.~Tao, and B.~Catanzaro, ``Image inpainting for irregular holes using partial convolutions,'' in {\em Proceedings of the European conference on computer vision (ECCV)}, pp.~85--100, 2018.

\bibitem{yu2019free}
J.~Yu, Z.~Lin, J.~Yang, X.~Shen, X.~Lu, and T.~S. Huang, ``Free-form image inpainting with gated convolution,'' in {\em Proceedings of the IEEE/CVF International Conference on Computer Vision}, pp.~4471--4480, 2019.

\bibitem{yi2020contextual}
Z.~Yi, Q.~Tang, S.~Azizi, D.~Jang, and Z.~Xu, ``Contextual residual aggregation for ultra high-resolution image inpainting,'' in {\em Proceedings of the IEEE/CVF conference on computer vision and pattern recognition}, pp.~7508--7517, 2020.

\bibitem{xie2019image}
C.~Xie, S.~Liu, C.~Li, M.-M. Cheng, W.~Zuo, X.~Liu, S.~Wen, and E.~Ding, ``Image inpainting with learnable bidirectional attention maps,'' in {\em Proceedings of the IEEE/CVF international conference on computer vision}, pp.~8858--8867, 2019.

\bibitem{ma2019coarse}
Y.~Ma, X.~Liu, S.~Bai, L.~Wang, D.~He, and A.~Liu, ``Coarse-to-fine image inpainting via region-wise convolutions and non-local correlation.,'' in {\em Ijcai}, pp.~3123--3129, 2019.

\bibitem{yu2018generative}
J.~Yu, Z.~Lin, J.~Yang, X.~Shen, X.~Lu, and T.~S. Huang, ``Generative image inpainting with contextual attention,'' in {\em Proceedings of the IEEE conference on computer vision and pattern recognition}, pp.~5505--5514, 2018.

\bibitem{lin2014microsoft}
T.-Y. Lin, M.~Maire, S.~Belongie, J.~Hays, P.~Perona, D.~Ramanan, P.~Doll{\'a}r, and C.~L. Zitnick, ``Microsoft coco: Common objects in context,'' in {\em Computer Vision--ECCV 2014: 13th European Conference, Zurich, Switzerland, September 6-12, 2014, Proceedings, Part V 13}, pp.~740--755, Springer, 2014.

\bibitem{lim2017geometric}
J.~H. Lim and J.~C. Ye, ``Geometric gan,'' {\em arXiv preprint arXiv:1705.02894}, 2017.

\bibitem{yang2017high}
C.~Yang, X.~Lu, Z.~Lin, E.~Shechtman, O.~Wang, and H.~Li, ``High-resolution image inpainting using multi-scale neural patch synthesis,'' in {\em Proceedings of the IEEE conference on computer vision and pattern recognition}, pp.~6721--6729, 2017.

\bibitem{xian2018texturegan}
W.~Xian, P.~Sangkloy, V.~Agrawal, A.~Raj, J.~Lu, C.~Fang, F.~Yu, and J.~Hays, ``Texturegan: Controlling deep image synthesis with texture patches,'' in {\em Proceedings of the IEEE conference on computer vision and pattern recognition}, pp.~8456--8465, 2018.

\bibitem{simonyan2014very}
K.~Simonyan and A.~Zisserman, ``Very deep convolutional networks for large-scale image recognition,'' {\em arXiv preprint arXiv:1409.1556}, 2014.

\bibitem{heder2022past}
M.~H{\'e}der, E.~Rig{\'o}, D.~Medgyesi, R.~Lovas, S.~Tenczer, A.~Farkas, M.~B. Em{\H{o}}di, J.~Kadlecsik, and P.~Kacsuk, ``The past, present and future of the elkh cloud,'' {\em INFORM{\'A}CI{\'O}S T{\'A}RSADALOM: T{\'A}RSADALOMTUDOM{\'A}NYI FOLY{\'O}IRAT}, vol.~22, no.~2, pp.~128--137, 2022.

\end{thebibliography}

\vspace{12pt}

\end{document}